# Inductive Policy Selection for First-Order MDPs

SungWook Yoon, Alan Fern, and Robert Givan
School of Electrical and Computer Engineering, Purdue University
West Lafayette, IN 47907

## Abstract

We select policies for large Markov Decision Processes (MDPs) with compact first-order representations. We find policies that generalize well as the number of objects in the domain grows, potentially without bound. Existing dynamic-programming approaches based on flat, propositional, or first-order representations either are impractical here or do not naturally scale as the number of objects grows without bound. We implement and evaluate an alternative approach that induces first-order policies using training data constructed by solving small problem instances using PGraphplan (Blum & Langford, 1999). Our policies are represented as ensembles of decision lists, using a taxonomic concept language. This approach extends the work of Martin and Geffner (2000) to stochastic domains, ensemble learning, and a wider variety of problems. Empirically, we find "good" policies for several stochastic first-order MDPs that are beyond the scope of previous approaches. We also discuss the application of this work to the relational reinforcement-learning problem.

## 1 Introduction

Many AI planning domains are naturally described in terms of objects and relations among objects—e.g., the blocks-world and logistics domains contain blocks, cars, trucks, and packages. Typically, such domains are compactly represented with first-order object quantification—e.g., "picking up any object results in holding that object."

Markov Decision Processes (MDPs) are a useful representation for stochastic planning domains. Research on MDPs, however, has dealt little with the issue of exploiting relational structure. Most existing algorithms for selecting control policies operate on either flat (Bellman, 1957; Howard, 1960; Puterman, 1994; Dean et al., 1995) or propositionally factored (Boutilier et al., 2000; Dean & Givan, 1997) representations. The size of a flat or propositional representation for a relational domain can be extremely large and is potentially infinite, and propositional algorithms are generally not polynomial in that size—rendering the associated algorithms impractical.

Recent MDP work uses a relationally factored value-function to carry out traditional dynamic programming methods (Boutilier et al., 2001). This technique powerfully exploits relational structure, but has two serious shortcomings addressed here. First, value-iteration approaches converge only after at least a number of iterations equal to the problem "solution length", as states have their value affected by rewards only at horizons sufficient to reach the rewards; however, the solution length can grow with the number of domain objects. Second, the size of the value-function representation can grow exponentially with the number of iterations as the state space may have exponentially many regions of different value.

Here, we examine planning problems that exhibit these phenomena when a value-iteration approach is applied.[1] Our approach does not compute a value function in large domains, but instead attempts to generalize good policies for domains with few objects to get a useful policy for domains with many objects. For example, patterns in the optimal solutions to five block blocks-world problems can be exploited in 50 block problems.

Policy construction by generalization from small problems was recently studied for deterministic problems by Khardon (1999) and Martin & Geffner (2000). Here, we extend that work to stochastic problems, widen the variety of domains considered, and consider a different taxonomic concept language for induced policies (i.e., a different language bias). We also add a heuristic concept selection technique and an ensemble learning method (bagging) and show substantial benefits from these extensions.

Our goals preclude guaranteeing an optimal or near-optimal policy—in many (even toy) planning domains, finding such a policy is NP-hard, or harder, and yet we would like to find useful policies in such domains.

This work raises the interesting question of whether pol-icy selection can be usefully improved by providing a "mostly optimal" policy—one that selects the optimal action at a high fraction of states. Intuitively, generalization from closely related, but solvable, problems, such as problems constructed by reducing the number of domain objects, may often produce policies that make good decisions in many states, but that make erroneous decisions in a (possibly) small fraction of states. Such policies can yield arbitrarily poor value functions—nevertheless, they represent a potentially rich source of information about an MDP's solution structure. In spite of this, most MDP research evaluates the utility of a policy based solely on its value function. We know of no work addressing policy selection when informed by such a "mostly optimal" policy. Our bagging technique combines a set of (hopefully) "mostly optimal" policies to get an "optimal" policy by voting, and is successful here.

---

[1] As an example problem, consider a blocks-world domain where the goal is to clear block **a**, where blocks have colors that affect the operators. While there is a very simple optimal policy, there are exponentially many uniform-value regions relative to the horizon, and solution length grows with domain size.

Another interesting problem raised by inductive policy selection is selection of "small" problem instances where the good policies are usefully related to good policies in large problems. While here we focus only on restricting the object domain size, construction of small instances by abstraction is also of interest. Generating useful abstractions automatically, and learning from the results of analyzing them, is a potential future direction.

Finally, this work is closely related to the relational reinforcement-learning problem, as we discuss in section 5.

## 2 First-Order Markov Decision Processes

In this work, we use a first-order stochastic planning language known as "first-order probabilistic STRIPS" (referred to from now on as PSTRIPS) that is the input language used by the stochastic planner PGraphplan (Blum & Langford, 1999), and is similar in expressive power and compactness to the situation-calculus–based language used by Boutilier et al. (2001). Our policy selection method is not tied to PSTRIPS, and could easily use a more general language—rather, we focus on this language because we use PGraphplan to generate training data from small problem instances. Our policy selection method applies to any MDP representation with a planner able to solve "small problem instances". (PGraphplan is such a planner for PSTRIPS; however, it propositionalizes the input problem, scaling poorly to large domains.)

### 2.1 First-Order Probabilistic STRIPS

In our variant, a PSTRIPS MDP is a tuple $<S,A,T,I>$, with each component described below.

**States.** Each MDP is associated with a finite set $S$ of predicate symbols that are interpreted as specifying properties of objects (single-arity predicates) and relations among objects (multi-arity predicates). Each state of an MDP is a first-order model of the associated predicates. That is, a state specifies a (finite) set of domain objects drawn from the natural numbers[2], and the truth value of each predicate application to those domain objects. For convenience, we assume each domain object (number) has a unique constant name and then represent states by listing the true ground facts. For example, the state

⟨ {a,b}, {on(a,b), clear(b), on-table(a)} ⟩

is a blocks-world state with exactly two blocks **a** and **b** in the domain, where **a** is on the table and **b** is on **a**. In general, there is no limit on the domain size of a state. The state space is therefore countably infinite, containing countably many states for each domain size. Below, in introducing goals, we give one restriction on the set $S$.

**Actions.** Our MDP actions are represented using a straightforward stochastic generalization of the commonly used deterministic STRIPS language (Fikes & Nilsson, 1971). Each MDP is associated with a finite set $A$ of action-type symbols, each of some specified arity. Given a state, each way of instantiating the action-type symbols with objects from the object domain in that state corresponds to an MDP action. For example, in the state shown above, the action **pick-up(a)** is an action of the single-arity type **pick-up**.

PSTRIPS compactly defines all actions of action-type $a$ via an action schema $T(a)$, using variables to abstract away from objects. An action schema has three parts:

1. *prototype*($T(a)$), which is an action-type symbol of arity $n$ applied to action variables $X_1, ..., X_n$.
2. *precondition*($T(a)$), a conjunction of MDP predicates applied to action variables from $X_1, ..., X_n$.
3. *outcomes*($T(a)$), a probability distribution (giving "occurrence probability") over a set of *possible outcomes*, each giving an add-list and a delete-list, each a set of MDP predicates applied to action variables.

The behavior of an action $a(o_1,...,o_n)$ in a state $q$ containing the $o_i$ is defined by first instantiating each $X_i$ with $o_i$ in the schema $T(a)$—this results in "ground" precondition and add/delete lists. Action $a(o_1,...,o_n)$ is *legal* in $q$ only if the ground precondition is true in $q$, and cannot be taken in $q$ otherwise. Each possible outcome of the action has a "possible next state" associated with it, when taken in state $q$—this is the state equal to $q$, but with any facts in the ground add (delete)-list added (deleted). If the action can be taken in state $q$, the next-state distribution is given by *outcomes*($T(a)$), with each possible outcome replaced by its possible next state, and other MDP states assigned probability zero. Deterministic STRIPS actions are just PSTRIPS actions with deterministic *outcomes*($T(a)$) distributions. Space precludes an example; see (Fern, URL).

Two factors often make it unnatural to capture a planning domain in PSTRIPS. First, PSTRIPS makes a fundamental assumption that the number of possible outcomes is not large—an assumption also present in the language of Boutilier et al. (2001). Thus, defining actions like "shuf-fle-cards" is clearly not feasible, requiring a possible outcome for each ordering of cards. Second, the possible outcomes are specified without quantification. Defining an action that knocks over a tower of arbitrary height is then difficult, since the most natural specification involves quantification. Despite these limitations, PSTRIPS still allows for challenging MDPs to be defined making it adequate for our initial investigation, and has an available, implemented planner for small problems (PGraphplan).

---

[2] Domains are finite subsets of number for simplicity, not necessity.

**Goal-Based Reward.** In order to use PGraphplan, we here consider only MDPs with goal-based reward structures—i.e., a set of goal states is specified as a conjunction of MDP predicates applied to objects and the objective is to expect to reach a goal state as quickly as possible. However, we note that our policy selection technique, in general, requires only a reward function language with a planner that can solve "small problem instances". Below, we describe how to specify goal states in our MDPs.

To facilitate generalization across different goals, we assume that the set $S$ of predicates is divided into "world predicates" and "goal predicates", with the two types of predicates in one-to-one correspondence. The world predicates are used to represent the current "world state"—in the blocks world, these might be **on(·,·)**, **ontable(·)**, and **clear(·)**. The goal predicates are used to represent the goals of the agent. We also restrict the PSTRIPS action definitions in $T$ to only add or delete world predicate facts. The systems of Khardon (1999) and Martin & Geffner (2000) also use world and goal predicates.

Conventionally, we name goal predicates by prepending a letter 'g' onto the corresponding world predicate—e.g., the goal predicate corresponding to **on(·,·)** is **gon(·,·)**. The MDP goal states are those states where, for every true goal predicate fact, the corresponding world fact is true. Thus, ⟨{a, b}, {on(a,b), clear(b), on-table(a), gclear(b)}⟩ is a goal state, but would not be so without **clear(b)**.

Our MDP state space has more states than truly intended. In the blocks world, there will be states where no block is on the table. Similarly, there will be states where the (unachievable) goal is to have every block on block **a**. Rather than attempt to give a language for axiomatizing the intended states and goals in the MDP, we instead assume that we are provided a problem-instance distribution $I$ over MDP states (which include the goal predicates) that describes the policy-selection problem of interest. In this work, we will describe this distribution in English, and implement it with a computer program that generates initial state/goal combinations from the distribution for each domain we study.[3] Our learning goal will be to find a policy that gives a low expected number of steps to a goal state from initial states drawn from the distribution $I$.

---

[3] This program must be able to condition the problem distribution on problem size, so that it can be used to generate problems of any given size.

### 2.2 Policy Selection

An MDP policy provides a mapping from states to actions—here, a mapping from first-order models to action types applied to domain objects from those models. Here, we focus on policy selection to minimize the expected number of actions to reach a goal state.

A primary goal of this work is to provide a policy selection method that scales well as the number of objects in an MDP grows. While it may be possible (or necessary) to re-plan for each different domain size, we focus here on finding good policies that apply to states involving any number of objects. As a simple example consider a deterministic blocks world MDP where the goal is to clear off a particular block. Clearly, a simple optimal policy applies to states with any number of blocks: "for any clear block above **a**, pick it up and put it on the table". Even in problems where finding the optimal policy is infeasible, there are sometimes (often?) "good" policies that generalize with the number of objects—e.g., there are well known "good" policies for (NP-hard) general blocks-world planning (Selman, 1994).

## 3 Learning Taxonomic Decision List Policies

### 3.1 Taxonomic Decision List Policies

Many useful rules for planning domains take the form "apply action type $a$ to any object in class $C$" (Martin & Geffner, 2000). For example, in the blocks world, "pick up any clear block that belongs on the table but is not on the table". Using a concept language for describing object classes, a class-based policy space has been shown to provide a useful learning bias for the deterministic blocks world (Martin & Geffner, 2000). In particular, such policies improve upon previous non-class-based blocks-world learning results (Khardon, 1999), without using the hand-engineered definitions that those results required.

With that motivation, we consider a policy space that is similar to the one used by Martin and Geffner. For historical reasons, our concept language is based upon taxonomic syntax (McAllester & Givan, 1993; McAllester, 1991), rather than on description logic.

#### 3.1.1 Taxonomic Syntax

Taxonomic syntax provides a language for writing class expressions, built from an MDP's predicate symbols, that describe sets of domain objects with properties of interest. Quantifier-free "taxonomic" concepts often require quantifiers to be expressed in first-order logic. For simplicity, we only consider predicates of arities one and two, which we call *primitive classes* and *relations*, respectively. Given a set of such predicates (the set $S$ defining the MDP states), class expressions are given by:

$C ::= C_0 \mid \textbf{a-thing} \mid \neg C \mid (R\ C) \mid C \cap C$

$R ::= R_0 \mid R^{-1} \mid R \cap R \mid R^*$

where $C$ is a class expression, $R$ is a relation expression,

$C_0$ is a primitive class, and $R_0$ is a primitive relation. Intuitively, the class expression $(R\ C)$ denotes the set of objects that are related through relation $R$ to some object in the set $C$. The expression $(R^*\ C)$ denotes the set of objects that are related through some "$R$ chain" to an object in $C$—this constructor is important for representing often-needed recursive concepts (e.g., the blocks above **a**).

Given an MDP state (i.e., a first-order interpretation) $q$ with domain $D$, the interpretation $C^q$ of a class expression $C$, relative to $q$, is a subset of $D$. A primitive class $C_0$ is interpreted as the set of objects for which predicate symbol $C_0$ is true in $q$. Likewise, a primitive relation $R_0$ is interpreted as the set of all object tuples for which the relation $R_0$ holds in $q$. The class-expression **a-thing** is interpreted to be $D$. For compound expressions,

$$(\neg C)^q = \{\, o \in D \mid o \notin C^q \,\}$$
$$(R\ C)^q = \{\, o \in D \mid \exists o' \in C^q, <o', o> \in R^q \,\}$$
$$(C_1 \cap C_2)^q = C_1^q \cap C_2^q$$
$$(R^*)^q = Id\ \cup$$
$$\{<o_1,o_k> \mid \exists o_2,\ldots,o_{k-1}\ \forall i\ <o_i,o_{i+1}> \in R^q\}$$
$$(R^{-1})^q = \{<o, o'> \mid <o', o> \in R^q\}$$
$$(R_1 \cap R_2)^q = R_1^q \cap R_2^q$$

where $C$, $C_1$, $C_2$ are class expressions, $R$, $R_1$, $R_2$ are relation expressions, and $Id$ is the identity relation. Some examples of useful blocks-world concepts, given the primitive classes **clear**, **gclear**, and **holding**, along with the primitive relations **on** and **gon**, are:

(**gon**$^{-1}$ **holding**), the block we want under the held block.

(**on*** (**on gclear**)) $\cap$ **clear**, clear blocks currently above blocks we want to make clear.

### 3.1.2 Decision List Policies

Like Martin and Geffner, we restrict to one argument action types $a_i$, and represent policies as decision lists:[4]

$C_1:a_1, C_2:a_2, \ldots, C_n:a_n$

where the $C_i$ are class expressions, and an expression $C_i:a_i$ is called a *rule*. Given an MDP state $q$, we say that a rule $R = C_i:a_i$ *suggests* an action $a_i(o)$ for $q$ if object $o$ is in $C_i^q$ and satisfies the preconditions of $a_i$ in $q$—the set of such actions is called *suggest*$(R, q)$. A single rule may suggest no action, or many actions of one type. We say a decision list *suggests* an action for state $q$ if a rule in the list suggests that action for $q$, and every previous rule suggests no action. Again, a decision list may suggest no action or many actions of one type. Each decision list $L$ for an MDP defines a policy $\pi[L]$ for that MDP—we assume an ordering on MDP actions, and if $L$ suggest no action for $q$, $\pi[L](q)$ is the least legal action in $q$; otherwise, $\pi[L](q)$ is the least action that $L$ suggests for $q$.

---

[4] Note that problems involving multiple-argument actions can be converted to 'equivalent' problems with only single argument actions. The resulting problems may be more difficult to solve, providing a practical motivation for special techniques for multiple-argument action types.

### 3.1.3 Policy-Space Restrictions

For effectiveness, we search through a restricted version of the policy space just described. The use of class and relational intersection is tightly controlled. Below we introduce "class-expression depth" to organize our search.

First, we introduce an abbreviation that we will "not expand" when measuring depth, to derive a useful language bias motivated by the classic AI planning principle of means-ends analysis (Newell & Simon, 1972). This principle suggests comparing the goal and current states, and selecting an action that maximally reduces the difference.

Leveraging the idea of comparing the goal and current states, we encourage our learner to use the intersection of a world predicate and corresponding goal predicate by treating such intersections as primitive predicates. Given a world predicate P (either a class or relation) and corresponding goal predicate $gP$, we write $cP$ (which we refer to as a "*comparison predicate*") to abbreviate $P \cap gP$. So, the fact **con(a,b)** abbreviates **(on$\cap$gon)(a,b)**, and indicates that block **a** is currently "correctly on" **b**. We consider a class expression to be "intersection-free" if the only uses of intersection occur inside comparison predicate abbreviations. This treatment of comparison predicates encourages our learner to use them aggressively.

We define the depth $d(C)$ of each intersection-free class expression $C$. The depth of **a-thing**, as well as any primitive or comparison class expression, is taken to be one. The depths $d(\neg C)$ and $d((R\ C))$ are both one plus $d(C)$, for any intersection-free relation expression $R$. So, **clear**, **gclear**, and **cclear** are all depth one, (**con*** **con-table**) has depth two (the set of blocks in well constructed towers), and (**gon** (**con*** **con-table**)) has depth three (blocks to be added to a currently well constructed tower).

To add intersection, define the set $C_{d,w}$ as the set of all classes formed by at most $w$ intersections, from depth $d$ intersection-free expressions. Excluding double negation and relation expressions that use either * or inverse twice, $C_{d,w}$ is finite for a given finite $S$. Our learning method uses a heuristic beam search to find useful concepts within $C_{d,w}$, where $d$ and $w$ are parameters of the algorithm.

### 3.2 A Greedy Learning Algorithm

We use a Rivest-style decision-list learning approach (Rivest, 1987)—an approach also taken by Khardon as well as Martin and Geffner. The primary difference between our technique and theirs is the method for selecting individual rules of the decision list. We use a greedy, heuristic search, while previous work used an exhaustive enumeration approach. This difference allows us to find rules that are more complex at the potential cost of failing to find some good, simple rules that enumeration might discover.

A training instance is a pair $<q, \alpha>$ where $q$ is a state and $\alpha$ is the set of actions that are desired in $q$. We say that a decision list $L$ *covers* a training instance $i = <q, \alpha>$ if $L$ suggests an action for $q$. We say that $L$ *correctly covers* $i$ if $L$ covers $i$ and the set of actions suggested by $L$ for $q$ is a subset of $\alpha$. Given a set of training instances, we will typically assume that the states of the instances all derive from the same MDP, and that the action sets contain only optimal actions for the corresponding states. Given these assumptions, if a decision list $L$ correctly covers a training instance, then $\pi[L]$ selects an optimal action for the corresponding state (under any ordering of the actions). This motivates searching for *consistent* decision-lists, those that correctly cover the training instances. The intent is to learn a decision list consistent with a sizable training-data set obtained by solving small-domain instances, and then apply that decision list to previously unseen MDP states with larger domains.

**Learning Lists of Rules.** Given a set of training instances we search for a consistent or nearly consistent decision list via an iterative set-covering approach. Decision-list rules $C:a$ are constructed one at a time and in order until the list covers (ideally, correctly covers) all of the training instances—we give pseudo-code for the algorithm in Algorithm 1. Initially, the decision list is the null list and does not cover any training instances. During each iteration, we search for a "high-quality" rule $C:a$, with quality measured relative to the set of currently uncovered training instances. The selected rule is appended to the current decision-list, and training instances covered by the new decision list, i.e., the ones newly covered by the new rule, are removed from the training data set. This process repeats until the list covers all of the training instances.[5] The success of this approach depends heavily on the function *Learn-Rule*, which selects a "good" rule relative to the uncovered training data—typically, a good rule is one that is consistent or nearly consistent with the training data, and also covers a significant number of instances.

**Learning Individual Rules.** The input to the learner is a set of training instances, along with depth and width parameters $d$ and $w$, and a beam width $b$ controlling the beam search described below. Currently, we focus on finding rules of the form $C:a$ with $C$ in $\mathbf{C}_{d,w}$ and $a$ an action-type symbol. We say a rule (correctly) covers a training instance when the decision-list containing only that rule (correctly) covers the instance—a rule is consistent with a set of training data if all of the instances it covers are correctly covered.

Algorithm 2 gives pseudo-code for our rule-learning algorithm, which uses two heuristics $H_1(\cdot)$ and $H_2(\cdot)$, described below, to rank candidate rules. First, for each action type $a$ we define a rule $R_a$, as follows: we conduct two beam searches, one with each heuristic function, to find two candidate rules using concepts from $\mathbf{C}_{d,w}$—we then choose the consistent rule if only one is consistent, and otherwise choose the $H_1$-selected rule. We have found this process to significantly improve results compared to using either heuristic alone. After rules $R_a$ have been defined for

---

**Learn-Decision-List ($F_0$, $d$, $w$, $b$)**
// training set $F_0$, concept depth $d$, width $w$, beam width $b$

$L \leftarrow$ NULL; $F \leftarrow F_0$;
while ($F$ is not empty)
    $C:a \leftarrow$ Learn-Rule($F$, $d$, $w$, $b$);
    $F \leftarrow F - \{f \in F \mid C:a \text{ covers } f\}$;
    $L \leftarrow$ extend-decision-list($L$, $C:a$); // end while

**Return:** $L$   // $L$ is a taxonomic decision list that covers $F$

Algorithm 1. Pseudo-code for Learn-Decision-List.

**Learn-Rule ($F$, $d$, $w$, $b$)**
// training set $F$, concept depth $d$, concept width $w$, beam width $b$

for each action type $a \in A$ // compute $C_a$ for each $a$
    $R_a \leftarrow$ Beam-Search($F$,$d$,$w$,$a$,$H_1$);
    if (not consistent?($R_a$, $F$))
        then $R' \leftarrow$ Beam-Search($F$,$d$,$w$,$a$,$H_2$);
            if (consistent?($R'$, $F$)) then $R_a \leftarrow R'$; // endfor

$X \leftarrow \{R_a \mid a \in A, \text{consistent?}(R_a, F)\}$
if ($X$ is empty) then $X \leftarrow \{R_a \mid a \in A\}$
**Return:** $\text{argmax}_{R \in X} H_1(R,F)$

Algorithm 2. Pseudo-code for Learn-Rule. Here, consistent?($R$,$F$) is true iff rule $R$ is consistent for instances $F$. $H_1()$ and $H_2()$ are the heuristic functions described in Section 3.2.

**Beam-Search ($F$, $d$, $w$, $b$, $a$, $H$)**
// training set $F$, concept depth $d$, concept width $w$, beam width $b$,
// action type $a$, heuristic function $H$

$B_0 \leftarrow \{ \text{a-thing} \}$;   $i \leftarrow 1$;   best $\leftarrow$ a-thing

while ((not consistent?(best:$a$, $F$)) &&
    ($i = 1$ || Hvalues($B_{i-1}$, $a$, $F$, $H$) !=
            Hvalues($B_{i-2}$, $a$, $F$, $H$))  )

    $G = B_{i-1} \cup \{ (C \cap C') \in \mathbf{C}_{d,w} \mid C \in B_{i-1}, C' \in \mathbf{C}_{d,1} \}$;
    $B_i \leftarrow$ beam-select($G$, $b$, $a$, $H$); // select $b$ best $H$ values
    best $\leftarrow \text{argmax}_{C \in B_i} H(C:a, F)$;
    $i \leftarrow i + 1$;                // end while

**Return:** best:$a$

Algorithm 3. Pseudo-code for Beam-Search. Here, the expression consistent?($R$,$F$) is true iff rule $R$ is consistent for instances $F$, Hvalues($B$, $a$, $F$, $H$) returns the set of heuristic values (measured by $H$) of members of $B$ when used in rules for action $a$ on instances in $F$; and beam-select($G$, $b$, $a$, $h$) selects the $b$ best concepts in $G$ with *different H* values (see footnote 6).

---

each type $a$, our rule-learning algorithm returns the rule $R_a$ with the highest $H_1$ value among those $R_a$ that are consistent, if any are consistent, or among all the $R_a$ otherwise.

Algorithm 3 gives pseudo-code for the beam search. To find $C_a:a$, given $a$, we generate a beam $B_0$, $B_1$, etc., of sets of class expressions from $\mathbf{C}_{d,w}$, repeatedly specializing expressions by intersecting them with other depth-$d$ class expressions, guided by the specified heuristic function. Search begins with only the most general concept, i.e., $B_0$ is the set {a-thing}. Search iteration $i$ produces a set $B_i$ that contains the $b$ class expressions with the highest *dif-*

---

[5] Every instance can be covered by using the **a-thing** class expression.

*ferent* heuristic values[6] among those in the following set

$$G = B_{i-1} \cup \{ (C \cap C') \in C_{d,w} \mid C \in B_{i-1}, C' \in C_{d,1} \}.$$

The sequence is terminated if the concept with the highest heuristic value in $B_i$ is consistent, or if there is no improvement in going from $B_{i-1}$ to $B_i$ (i.e., their elements yield the same set of heuristic values). We return the element of $B_i$ with the highest heuristic value.

**Heuristic Functions.** Heuristic functions $H_1$ and $H_2$ each take a rule $R = C : a$ and a set of instances $F$ as input, and return a pair of real numbers between zero and one, with

$$H_1(R,F) = <N_1(R,F), V(R,F)>, \text{ and}$$
$$H_2(R,F) = <N_2(R,F), V(R,F)>.$$

We take the heuristic values to be totally ordered, lexicographically. The value $V(R,F)$ is the fraction of the instances in $F$ covered by $R$, and each $N_i(R,F)$ measures rule consistency, as follows.

Define $F_a$ to be the set of all instances in $F$ where there is a legal action of type $a$. We evaluate $R$ by how well it suggests actions for the training instances in $F_a$. If $a$ is not a legal action for a state, then there is no decision to be made by $R$ at that state, so we ignore training instances outside of $F_a$.

To define $N_1(R,F)$, for each instance $f = <q, \alpha>$ of $F_a$, let $P(R,f)$ be the probability that a randomly selected action from *suggest*$(R,q)$ is in $\alpha$—when *suggest*$(R,q)$ is empty, so that no action is suggested, we take $P(R,f)$ to be zero if $\alpha$ contains any actions of type $a$, and one otherwise.[7] $N_1(R,F)$ is then the average value of $P(R,f)$ over all instances $f$ in $F_a$, but zero if $F_a$ is empty.

To define $N_2(R,F)$, let $X(R,F)$ be the number of examples in $F_a$ that $R$ covers incorrectly—$N_2(R,F)$ is equal to $1/(1+X(R,F))$. This heuristic is biased more heavily towards consistency than $N_1$.

### 3.3 Bagging

We intend our learner to learn patterns that select the optimal action at many states. Of course, this learner can be expected to make mistakes, given the inductive method of policy selection—we suggested above that this learner tries heuristically to produce a "mostly optimal" policy, selecting an optimal action at a high fraction of the states. One reason the policy may deviate from optimality is that practical constraints force our training sets to have limited size, so that some misleading patterns may appear, and our algorithm does nothing to control the standard machine learning problem of "overfitting" these patterns. We address these issues by using the ensemble method of "bootstrap aggregation", or "bagging" (Breiman, 1996). We note that other methods are available: overfitting can be controlled by larger training sets (possibly impractical) or regularization, and a mostly-optimal policy could potentially be improved by a heuristic search at run time.

In bagging, we generate several different training sets for the same MDP, and learn separate large-domain policies ("ensemble members") from each training set. We then combine these large-domain policies into one policy by voting. This approach addresses overfitting if the misleading patterns in the different training sets are independent, so that only a minority of the ensemble members are affected; the approach can be viewed as combining independent "mostly optimal" policies, assuming that the generalization errors made by each are independent.

It is usually the case that our learned policies make fatal mistakes in a small percentage of the trajectories used to test the policy. For example, a typical mistake we have observed in the blocks world is for a learned policy to **unstack** a block that is on top of a well-constructed tower. Such mistakes occur for example, when the last rule of a learned decision list is **a-thing : unstack** and a state with 'good towers' is encountered, where no previous rule suggests an action. When this happens, the next action selected by the policy is usually to stack the block back where it came from, resulting in an infinite loop. Typically, the rule suggesting the fatal action covers only a few examples, and most other ensemble members will not make the same mistake. Our experiments show bagging to be very effective at avoiding such actions.

Bagging requires additional parameters: an ensemble size $Z$ and a sample size $M$, and returns an ensemble (i.e., a set) of $Z$ decision lists found using our base learner on different training sets of size $M$. Specifically, given a set of training instances $F$, bagging proceeds as follows. First, we create $Z$ training sets $F_1, \ldots, F_Z$, all of size $M$ by randomly sampling $M$ training instances from $F$, with replacement. Next, we form an ensemble $E = \{L_1, \ldots, L_Z\}$, where $L_i$ is the decision list found using our base learner from Algorithm 1 applied to $F_i$. The policy $\pi[E]$ for the ensemble is defined using a simple vote among the ensemble members—so that $\pi[E](q)$, for state $q$, is equal to the action that is suggested for $q$ by the most members of $E$, breaking ties by selecting the least (legal) action.[8]

### 3.4 Training Example Generation

Our framework provides us with a distribution $I$ for generating initial states of a PSTRIPS MDP according to a distribution of interest. By conditioning this distribution on the object-domain size, we can control the complexity of the problem instances by varying the number of objects in the domain. It is important to note that the states generated by the program will not necessarily be representative of the states encountered later in full trajectories from generated initial states to generated goals. If not, learning from such training data is unlikely to produce a "good" policy at the un-represented states. To deal with this problem we augment the training data from the initial states provided by the problem generator with states occurring along "optimal" paths from those states to a goal. We use PGraphplan (Blum & Langford, 1999) to find such paths, and to find "optimal" actions for all the training data.

---

[6] Since many expressions in $C_{d,w}$ are equivalent, we must prevent the beam from "filling up" with semantically equivalent class expressions. Rather than deal with this problem via expensive equivalence testing we take an ad-hoc, but practically effective approach. We assume that class expressions do not coincidentally have the same heuristic value, so that ones that do must be equivalent. Thus, we construct beams whose members all have different heuristic values. We choose between class expressions with the same value by preferring smaller depths, then arbitrarily.

[7] $P(R,f)$ thus rewards rules for action type $a$ that suggest no action when no type $a$ action is optimal, but penalize them otherwise.

[8] Recall that a single ensemble member can suggest multiple actions of the same action type.

Table 1: Planning Domains

| | |
|---|---|
| **Blocks World 1 (BW$_1$).** One of the problems used to evaluate PGraphplan. World predicates are **on**(·,·), **on-table**(·), **clear**(·), and **holding**(·), with the standard blocks-world interpretations. Action types are **pickup** (...a block from the table), **put-down** (...the held block onto the table), **unstack** (...a block off a tower), **stack** (...held block onto a tower), **faststack** (move a block from the table to a tower[9]). Only **faststack** is stochastic, changing the state only with 0.8 probability. Problem size $p$ is a number of blocks, and initial and goal states of size $p$ are drawn uniformly with BWSTATES (Slaney, URL). We evaluate with $p$=6, $h$=20, $e$=80, $d$=3, $w$=12, $b$=5, and 20 block test problems. | **Logistics World 1 (LW$_1$).** Similar to that in (Boutilier et al., 2001). We have four object types **city**(·), **package**(·), **truck**(·), and **car**(·). Predicate **in**(·,·) used for packages in trucks/cars/ cities and for trucks/cars in cities. **selected**(·) predicate applies to trucks and cars, it is used to indicate which vehicle is involved in next action. Action types are **load**(pkg,vehicle), **unload**(pkg,vehicle), **drive**(vehicle, city), and **select**(vehicle). Only **drive** is stochastic, with success probability 0.9 for cars, 0.2 for trucks. Problem size is a vector giving the number of cities, cars, trucks, and packages. Distribution $I$ is given by uniformly distributing each vehicle among the cities, and each package among the vehicles and cities; with uniformly chosen goal cities for each package (and no other true goal facts). We evaluate with: $p$=<3 cities, 2 cars, 2 trucks, 3 pkgs>, $h$=20, $e$=160, $d$=4, $w$=12, $b$=5, and test problem size <5 cities, 7 cars, 7 trucks, 20 pkgs>. |
| **Blocks World 2 (BW$_2$).** As BW$_1$, except blocks are either **black**(·) or **gold**(·), and **faststack** success probability varies (0.8 black vs. 0.2 gold). Colors uniform at random. | |
| **Paint World 1 (PW$_1$).** As BW$_2$, except: **faststack** is removed, **stack** is now stochastic with the success probability varying with held block color, and new action **paint** 50% chance of changing held block color. Also, $p$=5, $h$=25 and $e$ = 100 (others unchanged). | **Logistic 2 (LW$_2$).** As LW$_1$, with a new predicate **rain**(·), and **drive** success probability is unchanged when no rain, but 0.8 for trucks in **rain** vs. 0.9 for cars in **rain**. **rain** is unchanging and uniformly random among cities. |
| **Paint World 2 (PW$_2$).** Same as PW$_1$ except success probability of **stack** also varies with destination color. | |

PGraphplan can be trivially adapted to accept a PSTRIPS MDP description, an initial state in that MDP, and a horizon time, and returns a contingent plan tree with maximum probability of reaching a goal state within the specified horizon time. This plan tree may *not* satisfy our objective function, which is to minimize the expected time to the goal. For example, if there is a long deterministic sequence of actions leading to the goal within the horizon time, that sequence of actions may be returned since it has a success probability of one. In such cases, however, there may be far better plans in terms of average plan length.

Rather than reject PGraphplan (which is one of the better publicly-available, open-source, probabilistic planners), we have chosen to use an ad-hoc technique that strongly encourages plans with short expected time to the goal. We simulate a discount factor (of 0.95) by modifying the original MDP to transition to a "dead" non-goal state with a fixed probability. Space precludes giving details here.

We note that an alternative here would be to use an MDP solver to return a complete policy for each small-domain MDP instance. We believe that explicit/flat MDP techniques will be impractical for this purpose, since even the small domains we are using here result in explicit MDPs that are near or beyond the limits of practicality for explicit techniques. A more promising alternative is to use solvers for propositionally factored (Boutilier et al., 2000; Guestrin et al., 2000) and relationally factored (Boutilier et al., 2001) MDPs. However, even small relational problems can give rise to relatively large proposition and action spaces, and yield complex and fragmented value functions. We also believe that it is both impractical and unnecessary to consider all of the information available in a complete small MDP policy.

To generate training data we specify a problem size $p$, a problem horizon $h$, and a trajectory count $t$. We sample $t$ initial states with problem size $p$, using the problem generating distribution $I$. For each of these initial states we then use PGraphplan with horizon $h$ to solve for trajectories to the goal by repeating the following steps either $h$ times or until a goal state is reached, whichever is first:

1. Beginning in the initial state use PGraphplan to generate an "optimal" contingent plan tree relative to the MDP, transformed to simulate discounting, as above.

2. Next, simulate the root action at the original MDP state, yielding a new "initial" MDP state.

The result is a sequence of states from some initial state provided by the problem generator to a goal state. For each state $s$ along the trajectory, we include the training example $<q,\alpha>$ where $\alpha$ is the set of all optimal actions in state $q$ according to PGraphplan[10]. We refer to the resulting training set with the random variable $train(I,p,t,h)$.

## 4 Experiments

### 4.1 Experimental Procedure

We evaluate our policy-selection approach on six PSTRIPS MDPs, described in Table 1, as follows. The parameters to our evaluation procedure are a PSTRIPS MDP definition $<S,A,T,I>$, a training-set problem size parameter[11] $p$, a training-set size $t$, training horizon $h$, a test set of 1000 initial states $Q$ drawn from $I$ conditioned on a problem size, an evaluation horizon $e$, and finally the concept depth $d$, concept width $w$, and beam width $b$ parameters required by our learning algorithm. For the ensemble learner, we use ensemble size 9, and sample 50 training instances for each ensemble member from a total training set of size 200.

---

[9] Since our system requires single-argument actions, we use a single-argument version of **faststack**, inducing the desired tower from the goal.

[10] We have trivially modified PGraphplan to return all optimal actions of the root rather than just one.

[11] The domain of this parameter varies—e.g., in logistics domains this may be a vector giving numbers of trucks, packages, etc.

A single trial of our evaluation proceeds as follows: draw a training set $F$ from $train(p,h,t,I)$, as described in Section 3.4. Next, let $L$ be the result of Learn-Decision-List$(F,d,w,b)$ (or corresponding ensemble hypothesis, in the case of bagging). Finally, for each initial state $q$ in the test set $Q$, run policy $\pi[L]$, starting at $q$, until either a goal state is reached, or more than $e$ actions have been executed. We return two numbers from each evaluation trial: the percentage $\phi$ of test problems from $Q$ where a goal was reached within the evaluation horizon $e$, which we call the *success probability*; and the average length $\psi$ of the trajectories that reached the goal. We run 40 evaluation trials for each MDP and report the average value of $\phi$ and $\psi$ over those trials.

Table 2. Evaluation Data

|     |        | $t=10$ | $T=50$ | $T=100$ | $t=200$ | $t=200+C$ | Bag  | Hand |
|-----|--------|--------|--------|---------|---------|-----------|------|------|
| $BW_1$ | $\phi$ | 0.67 | 0.83 | 0.82 | 0.91 | N/A | 1.0 | 1.0 |
|     | $\psi$ | 49.6 | 46.8 | 46.4 | 46.4 |     | 46.1 | 44.7 |
| $BW_2$ | $\phi$ | 0.49 | 0.82 | 0.86 | 0.89 | N/A | 0.98 | 1.0 |
|     | $\psi$ | 56.4 | 51.4 | 51.2 | 50.9 |     | 50.9 | 48.7 |
| $PW_1$ | $\phi$ | 0.41 | 0.88 | 0.89 | 0.91 | N/A | 0.99 | 1.0 |
|     | $\psi$ | 80.1 | 75.8 | 75.7 | 75.5 |     | 75.4 | 72.5 |
| $PW_2$ | $\phi$ | 0.09 | 0.43 | 0.5 | 0.42 | 0.58 | 0.97 | 1.0 |
|     | $\psi$ | 77.6 | 75.4 | 74.5 | 74.7 | 74.6 | 74.7 | 72.3 |
| $LW_1$ | $\phi$ | 0.66 | 0.82 | 0.78 | 0.93 | 0.99 | 0.96 | 1.0 |
|     | $\psi$ | 117 | 109 | 104 | 105 | 99.6 | 102 | 94.7 |
| $LW_2$ | $\phi$ | 0.41 | 0.76 | 0.85 | 0.85 | 0.94 | 0.96 | 1.0 |
|     | $\psi$ | 123 | 111 | 107 | 107 | 105 | 106 | 98.1 |

### 4.2 Results

**The Data.** Table 2 presents mean $\phi$ and $\psi$ values for the six domains for machine-learned single decision-list policies from four training-set sizes ($t=10$, 50, 100, 200), machine-learned ensemble policies (bag), and carefully hand-coded policies (hand).[12] One additional column ($t=200+C$) is explained below. The hand-coded policies are written in a richer language than our learned policies (e.g., allowing quantified taxonomic formulas), so the human coder can express concepts that the learner cannot.

**Varying Training-Set Size.** Both success probability $\phi$ and plan length $\psi$ generally improve with training set size—our method is turning training into improved policies. Even for the poor $t=10$, there is much improvement on the random policy ($\phi=0$). Additional training data may further improve $\phi$, as $\phi$ at $t=200$ still improves on $t=100$.

In contrast, the variation of $\psi$ at larger $t$ values is small. We speculate that larger training sets are needed primarily to avoid occasional "fatal" action choices, not to improve successful plan length.[13] Our bagging method provides an alternative attack on "fatal" choices, see Section 3.3

**Comparing to Previous Work.** To compare our technique with that of Martin and Geffner (2000), we evaluate our method in the same deterministic blocks world domain reported there. For a training set of 50 random five-block problems, Martin and Geffner (2000) report learning a policy achieving $\phi=0.722$ and $\psi=54.94$ when evaluated on 20 block problems. We ran 30 trials of the same experiment using our individual decision-list learner and 10 trials adding bagging (with ensemble size 7 and sample size 50). The policies learned by the individual decision list learner achieved $\phi=0.804$ and $\psi=55.4$, on average—improving on the success probability reported by Martin and Geffner. The average over all trials for bagging yielded $\phi=0.982$ and $\psi=56$—giving a further significant increase in success probability. It is unclear whether the improvement without bagging is due to our new heuristic learning method or our different underlying concept language. We expect that the use of bagging in conjunction with Martin and Geffner's decision-list learner would result in improvements similar to those seen here.

**Comparing to Hand-Coded Policies.** Humans win! The learned policies never outperform the hand-coded policies in either $\phi$ or $\psi$. Humans have no trouble constructing $\phi=1$ policies here, and work mainly on designing policies to reduce $\psi$ (typically by considering small problems).

The learner often finds rules that are similar or equivalent to parts of the human policies. Comparing the two, and designing (perhaps reasoning-based) methods to bridge the difference is a significant direction for future work.

**Bagging.** Bagging results for $t=200$ are a clear improvement over decision-list policies learned with the same amount of data, especially in $\phi$ (dramatically in PW$_2$). That $\phi$ improves much more than $\psi$ indicates that bagging is serving to filter out rare very "foolish" action choices that lead to failed policies. Although ensemble policies improve performance, a disadvantage is that they are difficult to analyze, either by hand or automated reasoning.

**Adding Concepts.** Our system uses a restricted concept language to facilitate effective learning—however, some useful concepts, typically requiring quantifiers, fall outside this language, and are exploited by humans in the hand-coded policies. It is trivial to enable our learner to exploit such concepts if they are provided as additional input by a human—simply treat the new concepts as primitive classes, and include them in constructed rules.

---

[12] We note that this small table summarizes an enormous amount of algorithm execution. For instance, each single decision-list policy entry corresponds to the generation of 40 training sets, each of size ten to twenty thousand, learning from these training sets, and then executing each of the resulting 40 policies from 1000 different start states to a significant problem-dependent horizon (or success).

[13] Recall, $\psi$ is the mean over successful trajectories only.

The column "$t=200+C$" reports three such experiments. For logistics, we added: "packages heading to the same city as a package in the selected vehicle" and "packages not currently at their goal". Adding these concepts allowed the learner to equal or beat the other learners, except the human. A similar experiment for $PW_2$ also shows a significant improvement, but significantly underperforms bagging.

## 5 Relational Reinforcement Learning

Our approach can be adapted for model-based, relational reinforcement learning (RRL). Exploration, along with some form of standard relational learning (e.g. Quinlan, 1990), can presumably be used to learn a relational transition model for the MDP (e.g., a PSTRIPS model for the actions). Learning the reward function is more complex: for an RRL problem to be plausibly solvable by any means, the reward function must either include some kind of "shaping" rewards (e.g., Mataric, 1994), in which case relational learning should be able to learn the function, or some access must be given to small problems (so "random wandering" can discover good policies). In previous RRL work, the latter case is typically assumed (Dzeroski et al., 2001), and we also take that approach here by assuming a problem generator, parameterized by problem size, for generating small instances.

Given means to learn the transition model and the reward model, the techniques in this paper can be applied to learn a policy that can then be greedily applied. We omit specifying exploration control for this method here.

Previous, Q-value–based, relational learners such as Q-RRL (Dzeroski et al., 2001) suffer from drawbacks like those described earlier for value-function–based approaches to relationally factored MDPs; these drawbacks can be avoided by using an inductive policy selection approach. This is the approach taken in P-RRL (also (Dzeroski et al., 2001)), where small problems are solved with Q-learning to provide policy-training data. In that work, learning was made practical by providing the learner with small problem instances in the early stages and then gradually increasing the problem size. We note that the experiments reported in that work involved simpler problems (e.g., placing all blocks on the table) than those we consider (e.g., building arbitrary towers). Q-RRL and P-RRL, both based on standard first-order logic syntax, also required the inclusion of human provided background knowledge in the form of predicate definitions (e.g., in the blocks world, the recursive predicate *above*). We show how to avoid providing background knowledge by choosing an appropriate policy language.

## 6 Conclusion

We have designed and empirically evaluated an inductive policy selection method for relationally factored MDPs. Exploiting solutions to small domain instances of an MDP, we learn policies that generalize well to larger domain sizes. Inspired by Martin and Geffner (2000), we utilize a policy language based on taxonomic syntax—this language allows for the compact representation of relationally factored policies, facilitating learning. We extend Martin and Geffner (2000) in a number of ways: considering stochastic MDPs, considering a wider variety of domains, introducing a heuristic learning method, improving performance using ensembles (i.e., bagging), and introducing a learning bias inspired by means-ends analysis.

Our method represents an alternative to structured dynamic programming (SDP) techniques for first-order MDPs. While first-order SDP techniques are a significant advance over flat or propositional techniques, they face fundamental difficulties when applied to the MDPs we consider here, due to complex value functions and solution lengths that grow with the number of domain objects.

## References

Bellman, R. (1957). *Dynamic Programming*. Princeton University Press.

Blum, A., & Langford, J. (1999). Probabilistic Planning in the Graphplan Framework. In *Proceedings of European Conference on Planning*.

Boutilier, C., Reiter, R., & Price, B. (2001). Symbolic Dynamic Programming for First-order MDPs. In *Proceedings IJCAI-01*.

Boutilier, C., Dearden, R., & Goldszmidt, M. (2000). Stochastic dynamic programming with factored representations. *Artificial Intelligence*, 121:49-107.

Boutilier, C., Dearden, R., & Hanks, S. (1999). Decision theoretic planning: Structural assumptions and computation leverage. *Journal of Artificial Intelligence Research*, 11:1–94.

Breiman, L. (1996). Bagging Predictors. *Machine Learning*, 24:123–140.

Dean, T., Kaelbling, L. P., Kirman, J., & Nicholson, A. (1995). Planning under time constraints in stochastic domains. Artificial Intelligence, 76.

Dean, T., & Givan, R. (1997). Model minimization in Markov decision processes. In *Proceedings AAAI-97*.

Dzeroski, S., Raedt, L., & Driessens, K. (2001). Relational Reinforcement Learning. *Machine Learning*,43:7–52.

Fern, A. URL: http://www.ece.purdue.edu/~givan/uai02.html.

Fikes, R., & Nilsson, N. (1971). STRIPS: A New Approach to the Application of Theorem Proving to Problem Solving. *Artificial Intelligence*, 2(3/4):189-208.

Guestrin, C., Koller, D., & Parr, R. (2001). Max-norm Projections for Factored MDPs. In *Proceedings IJCAI-01*.

Howard, R. (1960). *Dynamic Programming and Markov Processes*. MIT Press.

Khardon, R. (1999). Learning Action Strategies for Planning Domains *Artificial Intelligence*, 113:125–148.

Martin, M., & Geffner, H. (2000). Learning Generalized Policies in Planning Using Concept Languages. In *Proceedings KRR-00*.

Mataric, M. (1994). Reward Functions for Accelerated Learning. ICML-94.

McAllester, D. (1991). Observations on Cognitive Judgements. In *Proceedings of AAAI-91*.

McAllester, D., & Givan, R. (1993). Taxonomic Syntax for First Order Inference. *Journal of the ACM*, 40(2):246-283.

Newell, A., Simon, A. (1972). *Human Problem Solving*. Prentice Hall.

Puterman, M. (1994). *Markov Decision Processes—Discrete Stochastic Dynamic Programming*. John Wiley & Sons, Inc.

Quinlan, R. (1990). Learning Logical Definitions from Relations. *Machine Learning*, 5(3):239-266.

Rivest, R. (1987). Learning Decision Lists. *Machine Learning*, 2(3):229-246.

Selman, B. (1994). Near-Optimal Plans, Tractability, and Reactivity. In *Proceedings of KRR-94*.

Slaney, J. BWSTATES. URL:http://cslab.anu.edu.au/~jks/bwstates.html.